# Implementing BERT and fine-tuned RobertA to detect AI generated news by ChatGPT

**Zecong Wang[1], Jiaxi Cheng[2*], Chen Cui[3], and Chenhao Yu[3]**

[1]School of Software Technology, Zhejiang University, Hangzhou, China
[2]School of Computer Science and Engineering, University of New South Wales, Sydney, New South Wales, Australia
[3]Computer and Information Engineering College, Inner Mongolia Normal University, Hohhot, China

**\* Correspondence:**
Corresponding Author: Jiaxi Cheng
jiaxi.cheng@student.unsw.edu.au

## Abstract

The abundance of information on social media has increased the necessity of accurate real-time rumour detection. Manual techniques of identifying and verifying fake news generated by AI tools are impracticable and time-consuming given the enormous volume of information generated every day. This has sparked an increase in interest in creating automated systems to find fake news on the Internet. The studies in this research demonstrate that the BERT and RobertA models with fine-tuning had the best success in detecting AI generated news. With a score of 98%, tweaked RobertA in particular showed excellent precision. In conclusion, this study has shown that neural networks can be used to identify bogus news AI generation news created by ChatGPT. The RobertA and BERT models' excellent performance indicates that these models can play a critical role in the fight against misinformation.

## 1. Introduction

Many people, including software engineers and editors, have been using ChatGPT extensively during the past few months. It gives people a tool to work more quickly and produce more results in a given amount of time. However, the misuse of AI in the transmission of rumours has come under attention due to the proliferation of fake news produced by ChatGPT and the spread of misinformation [1]. Twitter, Facebook, Youtube, and Tiktok are the platforms that have drawn the most attention in this regard for spreading fake news. These AI-generated news stories move quickly and can reach a big audience, which has made them a fertile field for rumours. These rumours have the power to sway public opinion and decision-making, which could have catastrophic repercussions.

Despite the many advantages of AI chatbots, one drawback is the prevalence of false information and rumours, which can do a great deal of harm. On social media sites, these erroneous or deceptive pieces of information can spread swiftly, causing confusion and sometimes dangerous outcomes [2]. Particularly because of its rapid creation and capacity to reach a vast user base, ChatGPT has developed into a breeding ground for rumours. Because of this, it's critical to comprehend the traits of rumours that are produced by machines and how they spread in order to lessen their damaging effects.

Machine learning techniques have been widely employed by researchers to identify fake news on social media sites. For instance, Mendon et al. used wordbooks and machine learning techniques to track the origins and dissemination of rumours on Twitter after natural disasters [3]. In order to detect spam on social media, Jain et al. [4] combined a convolutional neural network with a long short-term memory network. The authors discovered that their strategy outperformed conventional machine learning techniques in the detection of spam. In order to identify erroneous or misleading information, several writers have used neural networks, including Gao et al. [5] and Kaliyar et al. [6], to recognise rumours on social media platforms. These approaches have demonstrated the effectiveness of using machine learning to detect rumors on social media platforms. However, there is still much to be learned about the spread of rumors on social media, including the characteristics of rumors that make them more likely to be shared and the impact they have on individuals and society [7].

BERT has been widely used as a pretrained transformer-based model in numerous natural language processing projects. RobertA is a state-of-the-art natural language processing model that has excelled on several challenges. In order to improve rumour detection, our goal was to take advantage of RobertA's capacity to precisely capture contextual and semantic features in the text input.

In this study, we used the improved RobertA and BERT to find rumours in the datasets. We used Twitter datasets containing rumours and news to assess the learning effectiveness of the models. We evaluated the effectiveness of the BERT and RobertA models in comparison to a number of baseline models, such as CNN, LSTM,

BiLSTM, and CNN-BiLSTM.

The essay is set up like follows: First, the Introduction introduces the fake news produced by ChatGPT and emphasises the need of comprehending this phenomena. Second, a review of the literature on the identification of rumours on social media platforms is provided. Third, word clouds and other visualisations of the datasets used in our investigation are shown in the Data Visualisation section. Fourth, the Model section describes the BERT and RobertA model used for detecting rumors in the datasets. Fifth, the Experiments section presents the results by figures and tables to evaluate the results of the several neural networks models and present the analysis. Finally, the conclusions section summarizes the whole paper and discusses the implications of our study.

## 2. Related Works

A lot of researchers are working to develop techniques for spotting incorrect or misleading material on social media platforms in the field of rumour detection. Numerous strategies have been put forth, including more contemporary ones like deep learning models as well as more established ones like traditional machine learning techniques like support vector machines (SVMs) and decision trees [8 – 10]. Researchers have been using natural language processing increasingly frequently in recent years [11]. For instance, Li et al. [12] used XGBoost and NLP to categorise rumours on the Chinese social media site Weibo. These techniques show how well NLP works for spotting rumours on social media networks. To comprehend how rumours spread on social media, other researchers used network analysis. For example, Wu et al. [13] proposed a graph-kernel based hybrid SVM classifier to study the spread of rumors on Sina Weibo. Their research provides valuable insights into the spread of rumors on social media and the ways in which they can be detected.

In the recent years, numerous researchers applied neural networks in rumor detection on social media [14-16]. These models have proven to be effective in identifying false or misleading information because of its abilities to capture complex patterns in the datasets. For example. Al-Sarem et al. [17]] implemented neural networks in rumor detection, which utilized a combination of a convolutional neural network (CNN) and a long short-term memory (LSTM) network to classify rumors on COVID-19. The authors found that their approach outperformed traditional machine learning methods in detecting rumors on the platform. Lotfi et al. [18] used GCN to classify rumors on Twitter, and Prakash et al. [19] used a neural network to classify rumors with attention mechanism. These studies demonstrate the effectiveness of neural networks in determination of rumors on social media.

As a transformer-based model, BERT has been applied in various of natural language processing areas, which contain sentiment analysis [20], machine translation, and question answering [21]. In recent years, BERT has also been used in the field of rumor detection on social media [22-23]. Some researchers implemented BERT to detect rumors on during the COVID-19 pandemic [24-25]. These studies demonstrate the effectiveness of BERT in detecting rumors on social media platforms.

In addition to BERT, RobertA is a transformer-based language model, similar to

BERT [26-27]. Similar to BERT, RobertA has achieved good results on multiple benchmarks and has been fine-tuned on various tasks for specific domains, including sentiment analysis [28], text classification [29], and rumor detection [30].

## 3. Datasets

The datasets we used contain a collection of real and fake news created by ChatGPT that were posted on the Internet, which is acquired from the website: https://www.kaggle.com/datasets/bjoernjostein/fake-news-data-set?resource=download. These datasets have been classified to real and fake news already by people, which are intended for research purposes, such as the development and evaluation of algorithms for rumour detection and veracity classification. The datasets include news and the labels. The news are organized into directories according to event, with subfolders for real and fake.

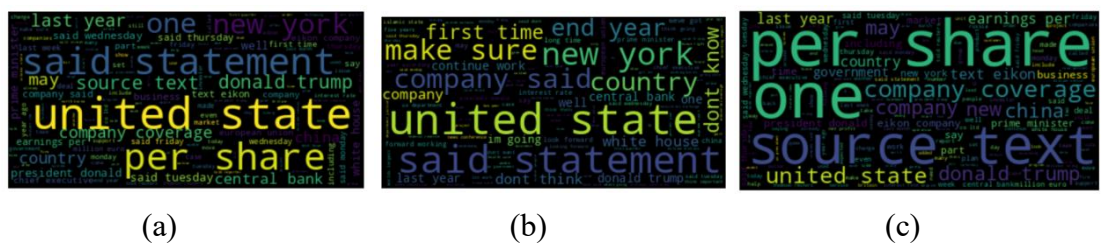

(a)　　　　　　　　　　(b)　　　　　　　　　　(c)

Figure 1. Word cloud of the datasets we used. (a) All the words in the datasets, (b) The fake (fake news), and (c) The real.

Figure 1 presents a word cloud for three datasets: (a) all words in the datasets, (b) rumors (fake news), and (c) real news. The size of each word in the word cloud represents its frequency in the respective dataset. The word cloud for all words (a) includes words that are commonly used across the datasets. The word cloud for rumors (b) includes words that are often associated with unverified or false information. The word cloud for non-rumors (c) includes words that are commonly used in factual, non-rumor news.

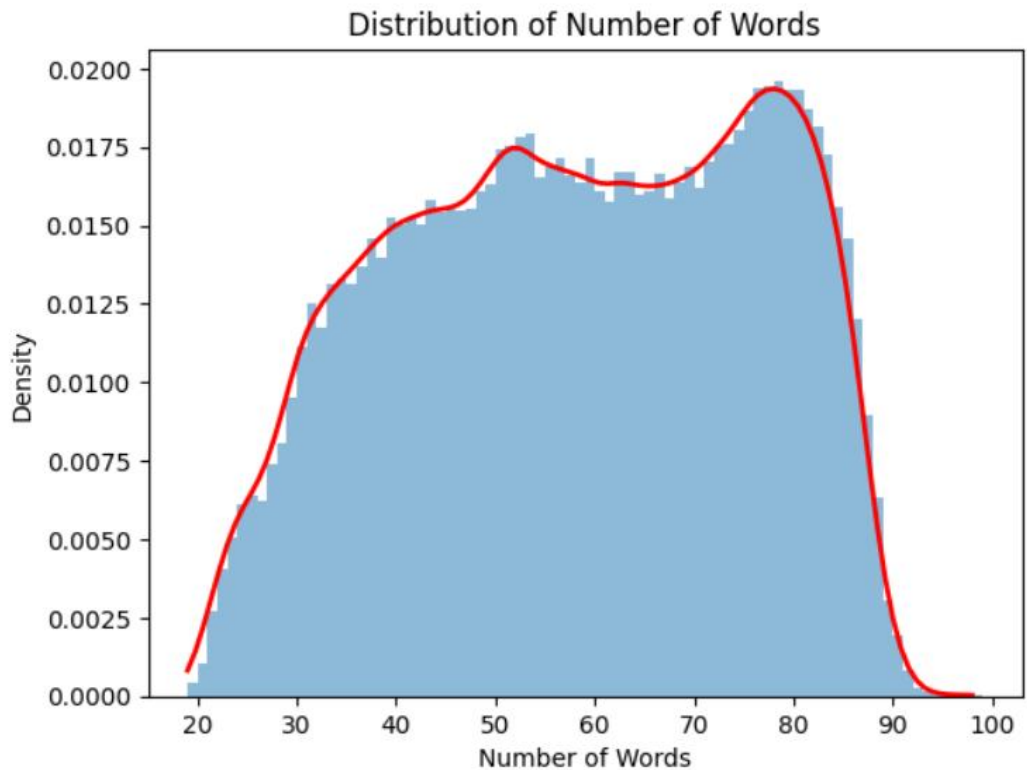

Figure 2. Distribution of Total Number of Words in each Tweets

Figure 2 shows the distribution of number of words in text of the datasets we used. It can be observed from the figure with a small number of outliers on both the lower and upper end. The median value is close to the peak, indicating that the majority of the texts in the datasets have a moderate number of words. It is also worth noting that the range of values for the number of words in the texts is relatively narrow, with the majority of texts falling within the 50 to 80 word range.

## 4. Models

In order to identify rumours, we will employ BERT and a modified RobertA model in this research. BERT is able to learn a variety of linguistic correlations and patterns since it is trained on a vast corpus of text data. It can comprehend a wide variety of linguistic patterns because it was trained on a diversified set of text material, including books, papers, and webpages [31–33]. BERT is also taught to comprehend the context of a statement, which is important for tasks like sentiment analysis and question answering.

In contrast, RobertA is a BERT version that uses a larger corpus and a different pre-training job. Even more data was used to train RobertA than was used to train BERT [26]. A comparable deep neural network architecture to BERT is used by Roberta, another transformer-based model, but with an extra task-specific pre-training. A separate pre-training task from the one used for BERT is utilised to train RobertA to anticipate the subsequent word in a phrase. Roberta employs the same self-attention technique as BERT and has a multi-layer transformer encoder as part of its architecture. This makes it possible for Roberta to comprehend the linguistic structure more thoroughly, which increases its effectiveness in some NLP.

The architecture of BERT and RobertA are each represented visually in Figure 3. The figure shows how BERT is constructed using a multi-layered transformer encoder architecture that includes a self-attention mechanism. The transformer encoder has a feed-forward neural network that converts the input into the output and an attention mechanism that can assist the model focus on particular input regions. The BERT's attention mechanism relies on self-attention, allowing the model to pay attention to multiple aspects of the input at once.

Roberta uses a similar architecture to BERT, but it uses a larger training corpus and a different pre-training task. A comparable deep neural network architecture to

BERT is used by Roberta, another transformer-based model, but with an extra task-specific pre-training. Predicting the subsequent word in a sentence is the additional pre-training challenge, which enables Roberta to get a deeper understanding of language structure. Roberta employs the same self-attention technique as BERT and has a multi-layer transformer encoder as part of its architecture. As a result, Roberta is able to comprehend the linguistic structure more thoroughly, which increases its effectiveness in specific NLP tasks.

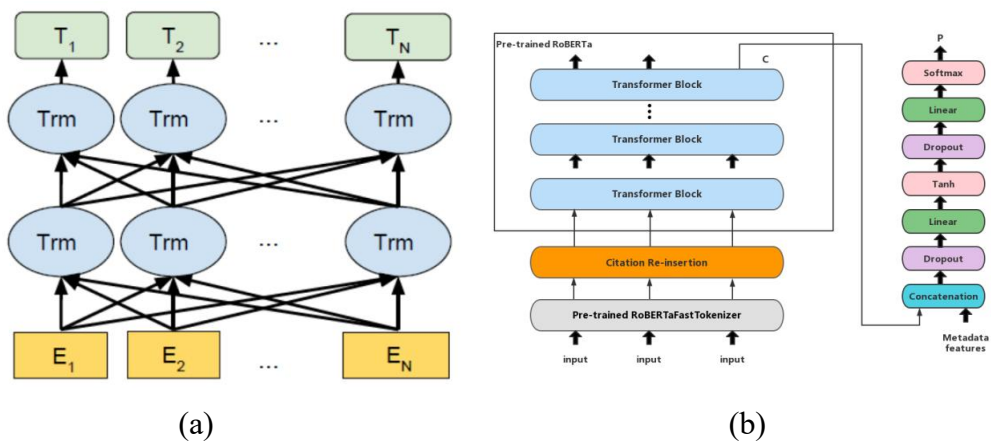

(a)          (b)

Figure 3 The (a) BERT and (b) RobertA transformer architecture

## 5. Experiments

IIn this section, we'll use a variety of neural network topologies to construct a Twitter rumour detection system. This system's objective is to categorise tweets as either true or fake rumours. To do this, we will examine a few traditional models and evaluate how well they do the task.

First, we'll incorporate CNN into our system for detecting rumours on Twitter. The ability of CNNs to extract features from images that are frequently employed in image classification tasks is one of their notable traits[34]. However, by considering the text as an image and utilising 1D convolutions, they can also be used for text classification tasks. Our CNN will have an input layer, many convolutional layers, and a fully connected output layer as part of its architecture.

Next, we will implement a LSTM structure [35] for our Twitter rumor detection system. In this paper, the architecture of our LSTM we used consisting of an input layer, multiple LSTM layers, and a fully connected output layer.

We will also implement a bidirectional LSTM (BiLSTM) [36] for our Twitter rumor detection system. BiLSTMs are similar to LSTMs, but they process the input sequence in both forward and backward directions. This feature permits it capture context from the past and the future. The architecture of our BiLSTM will also consist of an input layer, multiple BiLSTM layers, and a fully connected output layer.

Lastly, we will implement a CNN-LSTM [37] architecture for our Twitter rumor detection system. The CNN-LSTM architecture is a combination of the two previous models. It uses a CNN to extract features from the input text, and then an LSTM to process the extracted features. The architecture of our CNN-LSTM will consist of an input layer, multiple convolutional layers, multiple LSTM layers and a fully connected output layer.

Figure 4 plots the curves of the loss, accuracy, validation loss, and validation accuracy of CNN, LSTM, BiLSTM, and CNN-LSTM for Twitter rumor detection. As shown in the figure, all models performed well on the task with relatively low loss and high accuracy. However, the BiLSTM model had the best performance with regard to the validation loss and validation accuracy, indicating that it is the best model for the Twitter rumor detection system. The reason may be that the bidirectional structure of the BiLSTM allows the network to process the text in two directions: the past and the future, which is important for understanding the context of tweets. Furthermore, the BiLSTM model also had the lowest validation loss, indicating that it was able to generalize well to unseen data.

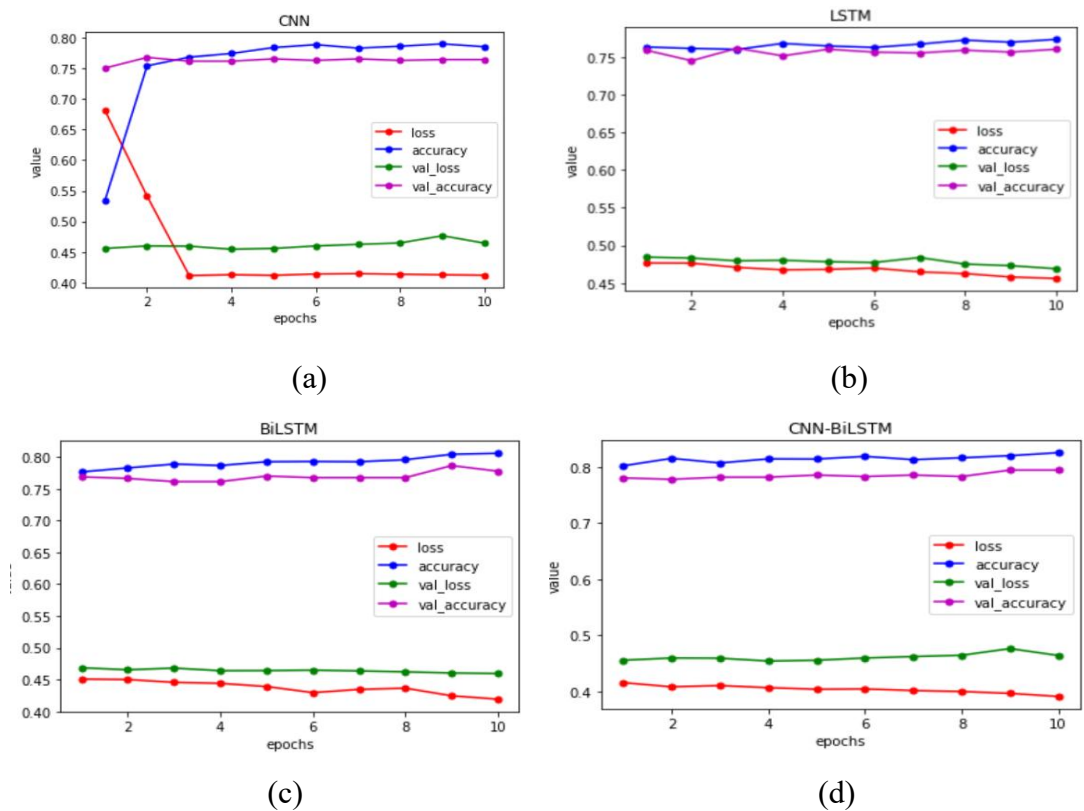

Figure 4 The performances of rumor detection by (a) CNN, (b) LSTM, (c) BiLSTM, and (d) CNN-BiLSTM

Figure 5 (a) and (b) displays the accuracy and loss of a BERT model when applied to our datasets, which we divide the datasets into training and validation parts. It can be observed from Figure 5(a) that the train accuracy is larger than 85%, indicating that the model has learned to accurately classify the examples in the train set. The validation accuracy is larger than 87%, which shows that BERT is able to generalize well to new examples. The validation accuracy is close to the train accuracy, which suggests that the model is not overfitting to the train data. The loss which shown in Figure 5(b) indicates the model learns to better classify the examples in the train set. It also proves that the model is also improving its performance on the validation set. The difference between the loss on the training and validation sets is relatively small, which is another indication that the model is not overfitting to the train data.

Figure 5 (c) and (d) represents the accuracy and loss of RobertA. It can be observed from the figure that the model achieved higher accuracy than BERT, which achieves 98%.It proves that RobertA has the perfect learning efficiency in rumor detection. The difference between the loss on the two datasets is relatively small, which suggests that the model is not overfitting to the train data.

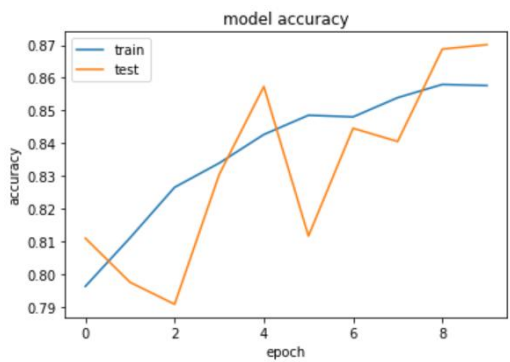
(a)

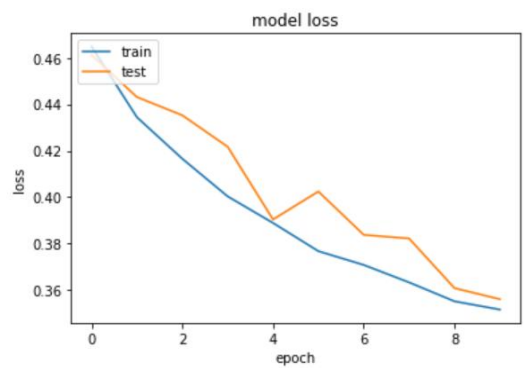
(b)

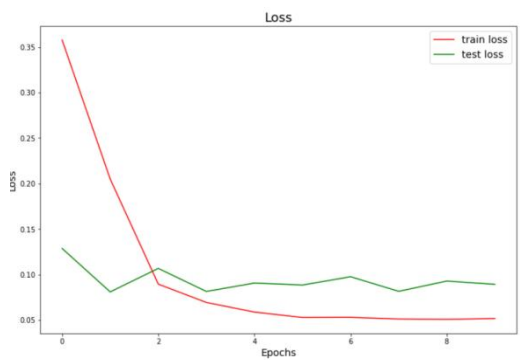

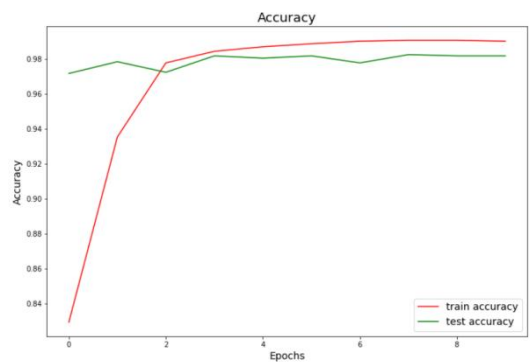

(c) (d)

Figure 5 The (a) accuracy and (b) loss of train (red line) and validation (green line) by BERT and the (c) accuracy and (d) loss of train (red line) and validation (green line) by fine-tuned RobertA

In the case of rumor detection, there are two classes: true (rumor) and false (real news). The confusion matrix in Figure 6 contains four values which indicates if the rumors have been correctly predicted. The Figure 6 indicated that both two model performed well on the task of rumor detection. The diagonal elements of the confusion matrix, which correspond to the true positives and true negatives, are much larger than the off-diagonal elements, which correspond to the false positives and false negatives. This indicates that the model correctly classified a large number of true and false rumors. Additionally, precision is a measure of the quality of the predictions made by the model. The BERT model shown in Figure 6 has a high precision, which means that it made high-quality predictions.

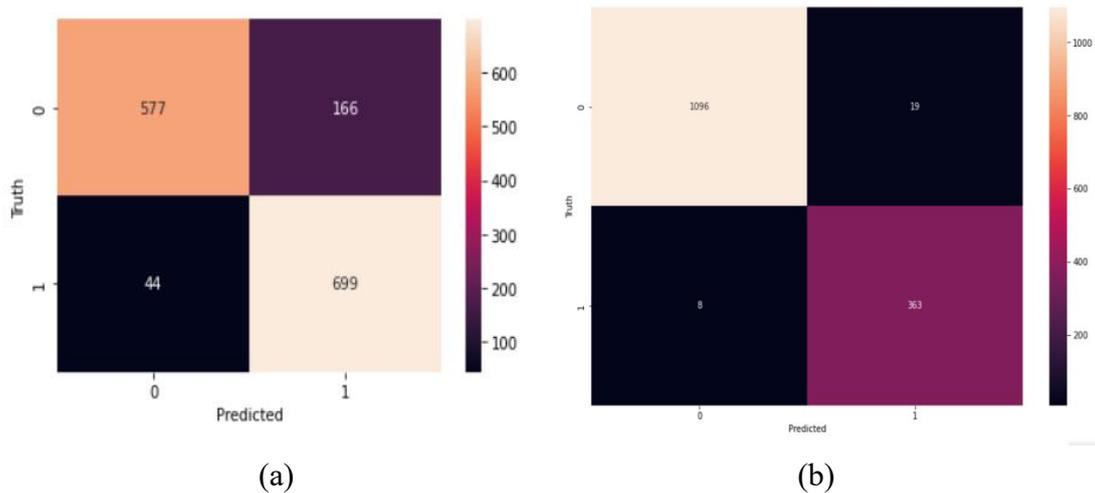

(a) (b)

Figure 6 The accuracy and loss of train and validation by (a) BERT and (b) fine-tuned RobertA

Table 1 offers a thorough overview of the BERT and fine-tuned RobertA models' classification performance on rumors and non-rumors. The table summarizes the precision, recall, F1 score, and support results for each model in the two classification categories. Precision reflects the accuracy of positive predictions, while recall indicates the proportion of positive cases correctly detected. The F1 score, which is the harmonic average of precision and recall, provides a unified metric that balances both values. Support represents the total number of observations in each category. As seen in the table, BERT has a precision of 0.93 and recall of 0.78 in classifying non-rumors, while RobertA has a precision and recall of 1 and 0.97, respectively. For the rumor classification, BERT has a precision of 0.81 and recall of 0.94, while RobertA has a precision of 0.92 and recall of 0.99. The F1 scores also reflect the overall performance of each model in the two classifications, with RobertA having higher scores than BERT in both cases. The support values in the table indicate that both models have been tested on a substantial number of observations for each classification category, providing reliable results.

Table 1 Classification report of BERT and RobertA model

| Classification | Model | Precision | Recall | F1 score | Support |
|---|---|---|---|---|---|
| Non-rumor | BERT | 0.93 | 0.78 | 0.85 | 743 |
| Rumor | BERT | 0.81 | 0.94 | 0.87 | 743 |
| Non-rumor | RobertA | 1 | 0.97 | 0.99 | 1115 |
| Rumor | RobertA | 0.92 | 0.99 | 0.96 | 371 |

## 6. Conclusions

The authors of this research intend to develop a neural network for rumour detection. In addition to more conventional models like LSTM, CNN, BiLSTM, and CNN-BiLSTM, they also examine a number of other models, such as BERT and RobertA. The results of our tests show that the BERT and fine-tuned RobertA models outperform the other models in terms of performance. Notably, the BERT model had an accuracy rate of more than 87%, while the improved RobertA model had an accuracy rate of 98%.

Modern language models have been successfully used by the authors to show how good rumour detection is. This is a substantial addition to the subject because earlier work frequently used simpler models to do this purpose. The use of BERT and RobertA demonstrates the potential for these models to be applied to other NLP tasks beyond sentiment analysis and question answering, where they have already shown great success.

The study's findings also offer insightful information on the advantages and disadvantages of each strategy. While the BERT model, for instance, performs admirably overall, its accuracy is still marginally worse to that of the improved RobertA model. This shows that there might be opportunity for improvement in the BERT model's tuning or finding different designs that are more effective at detecting rumours.

The outcomes also emphasise how crucial it is to choose a model while keeping in mind the particulars of the work at hand. Although LSTM, CNN, BiLSTM, and CNN-BiLSTM are well-suited for other NLP tasks, rumour detection is not one of their strong suits. This emphasises how crucial task-specific evaluation is when choosing a model, as opposed to merely depending on the performance of a model on a particular benchmark dataset.

In conclusion, by demonstrating the effectiveness of cutting-edge models for identifying rumours, the authors have had a substantial influence on the NLP field.

The findings highlight the importance of task specificity when choosing a model and provide insightful perspectives on the models' capabilities and constraints. The authors hope that their work will motivate further study in the area and encourage the use of cutting-edge models for NLP tasks other than sentiment analysis and question-answering.